\documentclass[journal,twoside,web]{IEEEtran}
\usepackage{tmi}
\usepackage{cite}
\usepackage{amsmath,amssymb,amsfonts}
\usepackage{algorithmic}
\usepackage{graphicx}
\usepackage{textcomp}

\usepackage{amsmath}
\usepackage{multirow}
\usepackage{multicol}
\usepackage{makecell}
\usepackage[utf8]{inputenc} 
\usepackage[T1]{fontenc}    
\usepackage{booktabs}       
\usepackage{amsfonts}       
\usepackage{nicefrac}       
\usepackage{microtype}      
\usepackage{hyperref}

\def\BibTeX{{\rm B\kern-.05em{\sc i\kern-.025em b}\kern-.08em
    T\kern-.1667em\lower.7ex\hbox{E}\kern-.125emX}}
\begin{document}
\title{Deep Mutual Learning among Partially Labeled Datasets for Multi-Organ Segmentation}

\author{Xiaoyu Liu, 
        Linhao Qu,
        Ziyue Xie, 
        Yonghong Shi, 
        and Zhijian Song
\vspace{-6mm}  
\thanks{This work was supported by the National Natural Science Foundation of China under Grant 82072021. (Corresponding authors: Yonghong Shi and Zhijian Song)}%
\thanks{All the authors are with Digital Medical Research Center, School of Basic Medical Science, Fudan University, Shanghai 200032, China. They are also with Shanghai Key Lab of Medical Image Computing and Computer Assisted Intervention, Shanghai 200032, China. (e-mail: \{liuxiaoyu21, zyxie22\}@m.fudan.edu.cn), \{lhqu20, yonghong.shi, zjsong\}@fudan.edu.cn).}
}

\maketitle

\begin{abstract}
The task of labeling multiple organs for segmentation is a complex and time-consuming  process, resulting in a scarcity of comprehensively labeled multi-organ datasets while the emergence of  numerous partially labeled datasets. Current methods are  inadequate in effectively utilizing the supervised information available from these datasets,  thereby impeding the progress in improving the segmentation accuracy. This paper proposes a two-stage multi-organ segmentation method based on mutual learning, aiming to improve multi-organ segmentation performance by complementing information among partially labeled datasets. In the first stage, each partial-organ segmentation model utilizes the non-overlapping organ labels from different datasets and the distinct organ features extracted by different models, introducing additional mutual difference learning to generate higher quality pseudo labels for unlabeled organs. In the second stage, each full-organ segmentation model is supervised by fully labeled datasets with pseudo labels and leverages true labels from other datasets, while dynamically sharing accurate features across different models, introducing additional mutual similarity learning to enhance multi-organ segmentation performance. Extensive experiments were conducted on nine datasets that included the head and neck, chest, abdomen, and pelvis. The results indicate that our method has achieved SOTA performance in segmentation tasks that rely on partial labels, and the ablation studies have thoroughly confirmed the efficacy of the mutual learning mechanism.
\end{abstract}

\begin{IEEEkeywords}
multi-organ segmentation, partially labeled, mutual learning.
\end{IEEEkeywords}

\section{Introduction}
\label{sec:introduction}
\IEEEPARstart{M}{ulti}-organ segmentation is crucial for various clinical tasks but remains a challenging problem in medical image processing. Although deep learning has advanced multi-organ segmentation models, training them typically requires the annotation of multiple organs as a prerequisite, which is both time-consuming and labor-intensive on a single medical image, such as CT or MRI scans \cite{25}. As a result, compared to natural image datasets, there is a limited number of public datasets available for training multi-organ segmentation model, making it difficult to meet the substantial data demands of deep learning. Additionally, the difficulty of annotating multiple organs has led many institutions to annotate only specific organs, resulting in numerous datasets with annotations for only some organs. For instance, the LITS dataset \cite{5} annotates only the liver and its tumor, while the KITS dataset \cite{6} focuses on kidney and kidney tumor. Leveraging these partially labeled datasets to develop models capable of segmenting multiple organs concurrently can reduce the annotation workload, improve segmentation accuracy, and meet urgent clinical needs.

\begin{figure}[t!]  
\centering
  \includegraphics[width=0.5\textwidth]{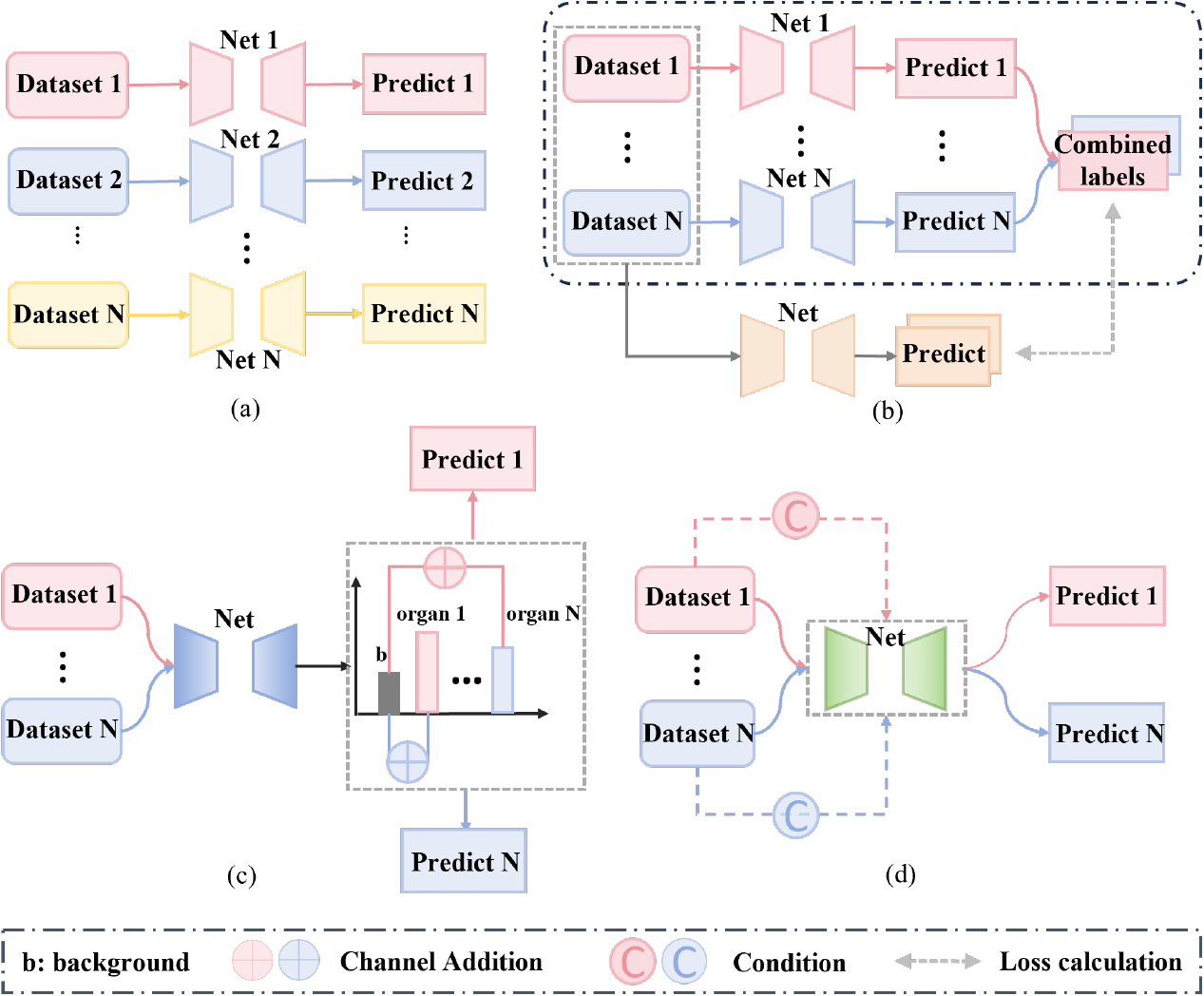}  
  \caption{Four types of methods to address partially labeled datasets for training multi-organ segmentation models. (a) multiple networks. (b) pseudo-labeling. (c) channel adjustment. (d) conditional information guidance.}
  \label{fig:1}
\end{figure}

A straightforward strategy to use these partially labeled datasets is to train a segmentation model independently for each dataset, and then combine the outputs of these models to obtain the final multi-organ segmentation result. This approach is known as \textit{multiple networks} and is shown in Fig.~\ref{fig:1} (a). Although simple, it has several obvious shortcomings: first, both the training and inference processes are time-consuming, while also necessitating substantial memory allocation for storing multiple models; second, during the inference stage, a challenge arises when integrating the outputs of individual models due to the voxel prediction conflicts, i.e., the prediction results for the same voxel may be not consistent from different models; and lastly, if datasets are trained independently with segmentation models, the prior information (such as size and position) contained between organs labeled in different datasets will not be effectively utilized, which makes it difficult to achieve optimal segmentation results \cite{7}.

The prevailing approach is to concurrently train a unified segmentation model using multiple datasets. Existing methods for training such models can be classified into three categories: \textit{Pseudo-Labeling}: As shown in Fig.~\ref{fig:1} (b), this approach initially trains models on individual datasets to segment specific organs, and then uses these models to generate pseudo labels for corresponding organs on other datasets. These combined labels are then used to train a unified multi-organ segmentation model. Research in this area focuses on enhancing the quality of pseudo labels \cite{8,9,24}.
\textit{Channel Adjustment}: Illustrated in Fig.~\ref{fig:1} (c) , this method employs a multi-channel output model. Due to the lack of labels for all channels, unlabeled channels are treated as background during loss calculation, which is called Target Adaptive Loss (TAL) \cite{10}. Shi et al. \cite{11} introduced additional marginal loss from other datasets to improve segmentation accuracy. Liu et al. \cite{7} initially trained a model with TAL and then iteratively refined it using self-training with pseudo labels.
\textit{Conditional Information Guidance}: Depicted in Fig.~\ref{fig:1} (d), this method integrates conditional information into the segmentation model, allowing the model to produce organ-specific segmentation result based on this information during inference. This conditional information is typically embedded into the network's final layers, guiding the network's output to correspond to the given condition \cite{12,13,14,30}.

However, existing methods do not fully integrate labeled organs from each dataset, leading to incomplete supervised information and limiting segmentation accuracy. Some methods also involve complex inference processes and voxel conflicts. Additionally, most current methods are based on abdominal datasets, and each dataset only labeled with a single organ. Other parts of the body (e.g., head and neck) also have partially labeled datasets, which may have more than one organ labeled, and the number of organs labeled varies from dataset to dataset, which poses a challenge to the model training and its generalization.

We've observed that there are interconnections among partially labeled datasets, and the models trained from each dataset are capable of learning from one another. Therefore, we introduce the concept of mutual learning for partial supervision. Mutual learning is a paradigm where multiple student networks collaborate, sharing knowledge to produce a more robust and adaptable network. \cite{15}. Recent studies have shown that multiple students learning together outperform individual learning \cite{16,17}. The application of mutual learning facilitates the exchange of the knowledge between datasets and models, thereby enhancing the segmentation accuracy through collaborative improvement.

\begin{figure}[t!]  
\centering
  \includegraphics[width=0.5\textwidth]{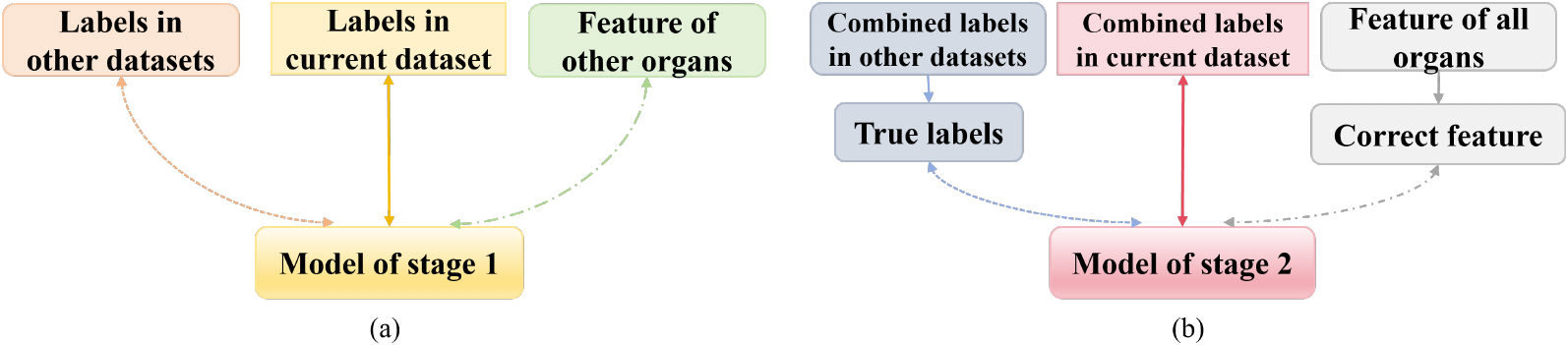}
  \caption{Main idea of our method. (a) each model in stage 1 is supervised by three kinds of information, including the labels in current dataset, the labels of the other datasets, and the features of other organs extracted by other models. (b) models in stage 2 are also supervised by three kinds of information, including the combined labels (with both the true labels and pseudo labels), the true labels of the other datasets, and the correct features extracted by the other models.}
  \label{fig:2}
\end{figure}

Our method consists of two stages, with each stage leveraging complementary information across datasets to enhance model performance. In the first stage, each student model is trained under supervision not only from the labels of the current dataset but also from the labels of other datasets and the features extracted by other models, as shown in Fig.~\ref{fig:2} (a). This stage enhances each model's ability to segment the current organ and improves the quality of pseudo labels, resulting in high-quality, fully labeled datasets with pseudo labels. In the second stage, each student model is supervised by the combined labels of the current dataset and the true labels from other datasets, while also being supervised by the correct features dynamically conveyed by other models in the latent space, as shown in Fig.~\ref{fig:2} (b), thus making full use of supervision information to improve the performance of multi-organ segmentation. The framework we proposed is capable of accommodating varying numbers of labeled organs across different datasets, including those contain multiple labeled organs.

The main contributions are as the following three aspects:

\begin{itemize}
    \item We introduce a two-stage mutual learning approach for partially labeled multi-organ segmentation. Each stage leverages complementary information across datasets to enhance supervised information, resulting in a model capable of accurately segmenting multiple organs simultaneously.
    
    \item Our mutual learning approach generalizes across various body regions, including the head and neck, chest, abdomen, and pelvis, and our method is adaptable to scenarios where multiple organs are annotated per dataset.
    
    \item The feasibility and effectiveness of our mutual learning mechanism are validated through experiments on datasets of multi-body regions, showcasing improved the accuracy and robustness of the segmentation.
\end{itemize}

\section{Related work}
\subsection{Multi-organ segmentation}
Accurate segmentation of multiple organs from the head and neck, chest and abdomen has always been a matter of great concern. In recent years, many effective methods have been proposed aiming to improve the performance of multi-organ segmentation. Some of these methods are from the perspective of network architecture design, such as the transformer \cite{18} and the two-stage method \cite{19}. Some utilise multi-view information \cite{20}; and some introduce effective modules, such as the attention module \cite{21} and dilated convolution \cite{22}, or design new loss functions \cite{23}, etc. The advancement of these methods effectively improves the performance of multi-organ segmentation. However, since it is very difficult to obtain a large number of fully labeled datasets, the training of models in the above studies is mostly restricted to a limited number of public datasets for multi-organ segmentation.

\begin{figure*}[t!]  
\centering
  \includegraphics[width=\textwidth]{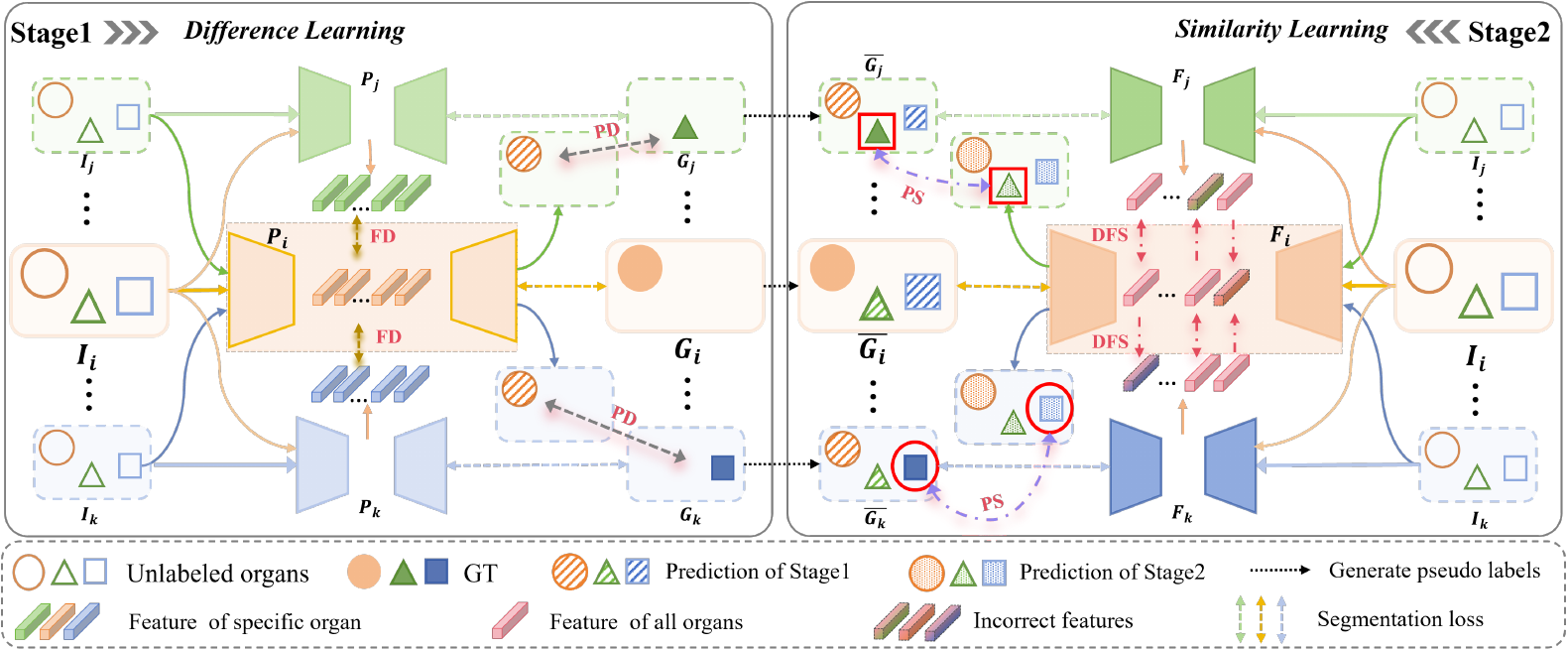} 
  \caption{The overall framework of our method. The first stage involves training multiple models to segment partial organs (from left to right). In addition to employing its own labels for supervised learning, each model participates in additional mutual Prediction Difference (PD) and Feature Difference (FD) learning. Then the trained models generate pseudo labels for other datasets, resulting in a combined labeled dataset. In the second stage, multiple models capable of segmenting all organs are trained (from right to left). During training, each model is supervised not only by the combined labels but also participates in additional mutual Prediction Similarity (PS) and Dynamic Feature Similarity (DFS) learning.}
  \label{fig:3}
\end{figure*}

\subsection{Partially labeled segmentation}
The substantial workload of annotating multiple organs in medical images (e.g., CT or MRI) has resulted in many datasets being only partially labeled, such as LITS \cite{5}, KITS \cite{6}, and PDDCA \cite{31}. To address this, various methods have been developed to train unified multi-organ segmentation models using these partially labeled datasets.

Existing methods are categorized into three primary categories:
\textit{Pseudo-Labeling}: These methods mostly depended on generating pseudo labels for unlabeled organs. Existing methods involve using pairs of networks for cooperative training to refine pseudo labels \cite{8} or leveraging label information from other datasets to enhance the quality of pseudo labels \cite{9}. However, the influence of pseudo labels and incomplete supervisory information make it difficult for multi-organ segmentation models to achieve significant further improvement.
\textit{Channel Adjustment}: These methods adjust the output channels of models to compute specialized TAL, by incorporating marginal losses from other datasets \cite{7,10,11}. This kind of methods primarily deal with each dataset individually, and when testing, they also need to adjust the channels of the output results in order to obtain the segmentation results of the corresponding organs.
\textit{Conditional Information Guidance}: These methods \cite{12,13,14,30} incorporate conditional information to guide segmentation. They sequentially infer organ-specific results, potentially leading to voxel prediction conflicts, and require extensive training and inference time.

\subsection{Mutual Learning}
Unlike previous studies, our method employs mutual learning to address the partially labeled multi-organ segmentation problem. Mutual learning involves multiple student networks that guide and learn from each other during training, leading to the development of a more robust and widely applicable network. This approach deviates from traditional distillation method that rely on a teacher-student model, instead, mutual learning is emphasized where students exclusively instruct one another. Zhang et al. \cite{15} first proposed the concept of mutual learning, demonstrating that peer teaching results in better performance than isolated supervised learning. Fang et al. C. \cite{16} extended this concept to medical image segmentation, employing dual segmentation models to reduce noise in imperfect annotations by providing clean training data to each other. Zhu et al. \cite{17} introduced an online mutual learning strategy where a CNN and a ViT collaborate, leveraging their complementary strengths and compensating for their respective limitations. In our study, each model learns new features and patterns from the other during training. This reciprocal learning enriches the supervisory signals and enhances segmentation accuracy, thereby resulting in a more effective model for multi-organ segmentation.

\section{Method}

\subsection{Overview}
Given \( N \) datasets \(\left\{D_i\right\}_{i=1}^N\), the \( i \)-th dataset \( D_i = \left\{I_i, G_i, O^i\right\}\), where \( I_i = \left\{I_{ij}\right\}_{j=1}^{N_i} \) and \( G_i = \left\{G_{ij}\right\}_{j=1}^{N_i} \), \( I_{ij} \) represents the \( j \)-th image in \( D_i \), and \( G_{ij} \) denotes the corresponding Ground Truth (GT). \( N_i \) indicates the number of images in the \( i \)-th dataset. $O^i$ represents the set of annotated organs in \( D_i \). $G_{i}(O^i)$ represents the GT of organ $O^i$ in \( G_i \). The union of labeled organs across all datasets is $O$. Clearly, \(O^i \subseteq O\). For any two datasets \( D_m \) and \( D_n \), their annotated organs do not overlap, i.e., $ O^m \cap O^n = \emptyset$. The union of all annotated organs across datasets equals the total set of annotated organs, i.e., \(\bigcup_{i=1}^{N}O^i = O\). Our goal is to train a model \( F \) using these datasets. When given an unlabeled input image, \( F \) will get segmentation results for all target organs. 

Fig.~\ref{fig:3} illustrates the proposed method, which involves two stages. The goal of first stage is to obtain multiple Partial-Organ Segmentation (POS) models, where each POS model, denoted as \( P_i \), acts as a student model to segment different organs. Beyond the supervised learning using its own labels, each \( P_i \) also engages in additional difference mutual learning, which includes both Prediction Difference (PD) and Feature Difference (FD). The trained models then generate pseudo labels for other datasets, resulting in combined labeled datasets. The goal of second stage is to train multiple Full-Organ Segmentation (FOS) models using the fully labeled datasets with pseudo labels. Each FOS model, denoted as \( F_i \), acts as a student model to segment all target organs. Similarly, additional Prediction Similarity (PS) and Dynamic Feature Similarity (DFS) mutual learning are introduced during training. Detailed information are as follows.

\subsection{Difference Mutual Learning}

We trained \( N \) POS models \(\{P_i\}_{i=1}^N\) on \( N \) partially labeled datasets \(\{D_i\}_{i=1}^N\), where each \( P_i \) model is considered a student model. In addition to employing its own labels for supervised learning, each \( P_i \) participates in addtional mutual learning processes, encompassing prediction and feature difference learning:

\subsubsection{Prediction Difference Learning}
As illustrated in Fig.~\ref{fig:3}, taking dataset \( D_i \) as an example, it contains \( I_i \) (Input), \( G_i \) (GT) and \( O^i \) (labeled organ set). The main segmentation loss \( L_i^1 \) for \( P_i \) is calculated between the network's predicted results \(\text{pre}_{i\_i}^1\) and \( G_i \), as shown in Equation (2). However, when other datasets, such as \( D_j \) and \( D_k \), are input into \( P_i \), predictions for organ \( O^i \) are generated. Since the labels of organs are mutually exclusive, the predictions for organ \( O^i \) should not overlap with the GT in \( D_j \) and \( D_k \). Hence, we propose a \textbf{Prediction Difference (PD) Loss}  \( L_i^{1\_l} \), where a larger loss indicates that the predictions of \( P_i \) on other datasets do not overlap with the annotated organs, implying better segmentation performance, as shown in Equation (3).

\subsubsection{Feature Difference Learning}
Simultaneously, to further distinguish the segmentation capabilities of different models, we introduced a \textbf{Feature Difference (FD) Loss}. Specifically, when the image from \( D_i \) is input into different student models \( P_i \), \( P_j \), and \( P_k \), the highest-level semantic features extracted by the encoder are \( f_{i\_i}^1 \), \( f_{j\_i}^1 \), and \( f_{k\_i}^1 \). The greater the difference between \( f_{i\_i}^1 \) and \( f_{j\_i}^1 \) and \( f_{k\_i}^1 \), the larger the difference in features extracted by different models. Therefore, based on \( L_i^1 \) and \( L_i^{1\_l} \), we incorporate a feature difference mutual learning loss \( L_i^{1\_f} \), as shown in Equation (4).

The introduction of difference learning not only enhances the segmentation accuracy of \( P_i \) but also enables \( P_i \) to perceive the presence of unannotated organs, thereby improving the quality of pseudo labels generated for these organs on other datasets. This method is extended to all training datasets, where PD loss and FD loss is calculated between any two datasets. The specific loss function calculations are as follows:

\jot=2pt
\begin{equation}
L^1 = \sum_{i=1}^N L_i^1 - \lambda_l \left( L_i^{1\_l} \right) - \lambda_f \left( L_i^{1\_f} \right)
\end{equation}
\begin{equation}
L_i^1=L_{D_i}\left(\operatorname{pre}_{i_{-} i}^1(O^i), G_i(O^i)\right)
\end{equation}
\begin{equation}
L_i^{1\_l}=\frac{1}{N-1} \sum_{j, j \neq i} L_{D_j}\left(\operatorname{pre}_{i_{-} j}^1(O^i), G_j(O^j)\right)
\end{equation}
\begin{equation}
L_{i}^{\text {1\_f}}=\frac{1}{N-1} \sum_{j, j \neq i} \cos \left(f_{i\_i}^1, f_{j\_i}^1\right)
\end{equation}

Among them, $L_i^1$ represents the segmentation loss, $L_i^{1\_l}$ represents the PD loss, and $L_i^{1\_f}$ represents the FD loss. $pre_{i\_i}^1$ and $pre_{i\_j}^1$ denote the predictions of $P_i$ applied to $D_i$ and $D_j$, respectively. $G_i$ and $G_j$ represent the labels for $D_i$ and $D_j$, and $f_{i\_i}^1$ and $f_{i\_j}^1$ indicate the features extracted by $P_i$ and $P_j$ on dataset $D_i$. The parameters $\lambda_{l}$ and $\lambda_{f}$ represent the hyper-parameters for the first stage.

\subsection{Generating pseudo labels}

After completing the training of \(\left\{D_i\right\}_{i=1}^N\) in the first stage, pseudo labels are generated on other datasets, resulting in fully labeled datasets containing pseudo labels. When generating pseudo labels, if there is an overlap with the true labels of the current dataset, the true labels are prioritized. Upon obtaining the fully annotated dataset, the second stage of training begins.

\subsection{Similarity Mutual Learning}

In the second stage, we obtain the fully annotated datasets with pseudo labels, denoted as \(\left\{\bar{D}_i\right\}_{i=1}^N\), the \( i \)-th dataset \( \bar{D}_i = \left\{I_i, \bar{G}_i, \bar{O}^i\right\}\). The labels $\bar{G}_i$ include both the true labels $G_i$ and pseudo labels $\hat{G}i$ generated by other models. i.e., \( \bar{G}_i = \left\{G_i, \hat{G}^i\right\}\). The organ sets are denoted as $O^i$ and $\hat{O}^i$, respectively, i.e., \( \bar{O}_i = \left\{O_i, \hat{O}^i\right\}\), with $\hat{O}^i = \bigcap_{j \neq i}^{N} O^j$. Despite efforts to enhance the quality of pseudo labels in the first stage, inaccurate pseudo labels still negatively impact model training. To fully exploit the supervision information provided by the label characteristics of each dataset, we introduce similarity learning among multiple FOS models \(\{F_i\}_{i=1}^N\).

\subsubsection{Prediction Similarity Learning}
Specifically, taking dataset $\bar{D}_i$ as an example, the segmentation loss $L_i^2$ of model $F_i$ is computed from the network output $pre_{i\_i}^2$ and the labels $\bar{G}_i$. The presence of pseudo labels provides additional supervision but can also misguide the model. We note that dataset $\bar{D}_j$ contains the true labels for organs $O^j$. When $F_i$ is applied to dataset $\bar{D}_j$, it yields prediction results for organs $O^j$. The loss computed between these results and the true labels of $O^j$ can enhance the performance of $F_i$ in segmenting organs $O^j$. This loss is termed the \textbf{Prediction Similarity (PS) Loss}, with the main idea being the use of true labels from other datasets for supervision.

\subsubsection{Dynamic Feature Similarity Learning}
In addition to PS loss, we introduce Feature Similarity (FS) Loss similar to the first stage. However, unlike the first stage, each $F_i$ in this stage can segment all organs comprehensively. The highest-level semantic features $f_{i\_i}^2$ extracted by each student model includes features for all organs. Directly computing mutual learning loss could lead to inaccuracies, as features are extracted from models trained on pseudo labels. Given that $F_i$ and $F_j$ have different segmentation capabilities for different organs, we dynamically transfer the correct features between the two student models in the latent space to benefit each other. However, determining the direction of knowledge transfer during training is a challenging problem. Inspired by the mutual learning between CNN and Transformer \cite{17}, we propose managing the direction of knowledge transfer by combining prediction results with true labels, as follows:

Given the features $f_{i\_i}^2$ extracted by $F_i$ from $I_i$ and $f_{j\_i}^2$ extracted by $F_j$ from $I_i$, we first compute the cosine similarity $S_{(i,j)} = \cos(f_{i\_i}^2, f_{j\_i}^2)$. Then, we quantify the reliability of the knowledge between the two students using the cross-entropy loss between prediction and true labels. Specifically, we use $M_{(i,j)} \in \{0, 1\}$ to represent the direction of feature transfer. We calculate the prediction results \(\text{pre}_{i\_i}^2\) of $F_i$ on $I_i$ and \(\text{pre}_{i\_j}^2\) of $F_j$ on $I_i$, respectively, and then compute the cross-entropy loss with true labels $G_i$ to obtain $C_{i\_i}$ and $C_{i\_j}$. If $C_{i\_i}$ is larger than $C_{i\_j}$, it indicates that $F_j$ is more accurate than $F_i$, so $M_{(i,j)} = 0$, meaning the feature is transferred from $F_j$ to $F_i$. Otherwise, $M_{(i,j)} = 1$. Through this approach, $F_i$ and $F_j$ can exchange reliable knowledge, enabling the correct transfer of features, which is called \textbf{Dynamic Feature Similarity (DFS) Loss}.

The similarity learning allows each model to fully utilize supervisory information, including true labels, pseudo labels, and correct features. This method is generalized to all training datasets, with similarity loss computed between any datasets. The specific loss function calculation is as follows:

\jot=3pt
\begin{equation}
L^2=\sum_{i=1}^N L_i^2+\beta_{l}\left(L_i^{2\_l}\right)+\beta_{f}\left(L_i^{2 \_f}\right)
\end{equation}
\begin{equation}
L_i^2=L_{\bar{D}_i}\left(\operatorname{pre}_{i\_i}^2(\bar{O}^i), \bar{G}_i(\bar{O}^i)\right)
\end{equation}
\begin{equation}
L_i^{2\_l}=\frac{1}{N-1} \sum_{j, j \neq i}\left(L_{\bar{D}_i}\left(\operatorname{pre}_{i_{-} j}^2(O^j), \bar{G}_j(O^j)\right)\right)
\end{equation}
\begin{equation}
L_i^{2\_f}=\frac{1}{N-1} \sum_{j, j \neq i}\left(1-M_{(i, j)}\right) S_{(i, j)}
\end{equation}

In this context, $L_i^2$ represents the primary segmentation loss, $L_i^{2\_l}$ represents the Prediction Similarity loss, and $L_i^{2\_f}$ represents the dynamic feature mutual learning loss. $pre_{i\_i}^2$ denotes the prediction results of $F_i$ applied to $\bar{G}_i$, The parameters $\beta_{l}$ and $\beta_{f}$ denote the hyper-parameters for the second stage.

\subsection{Inference Stage}

During the inference stage, the multi-organ segmentation model $F_i$ that performed best in the second stage is used as the final model $F$ for inference.

\section{Experiments}
\subsection{Dataset}
In this experiment, we established four tasks using nine public datasets, covering the head and neck, chest, abdomen, and pelvis. The specific datasets used for each region are as follows:

\textit{Head and Neck}: We used the PDDCA\cite{12} and StructSeg (https://structseg2019.grand-challenge.org/Dataset/) datasets. PDDCA contains 48 cases with 9 labeled organs; we selected the organ of brainstem, left optic nerve, and right optic nerve. StructSeg includes 60 cases with 22 labeled organs; we selected the organ of the chiasm, left parotid gland, right parotid gland, and mandible.

\textit{Chest}: We used the SegThor\cite{26} and StructSeg datasets. SegThor has 40 cases with 4 labeled organs (heart, aorta, trachea, and esophagus); we selected the heart and trachea. StructSeg includes 60 cases with 6 labeled organs (left lung, right lung, spinal cord, esophagus, heart, and trachea); we selected the left lung, right lung, and esophagus.

\textit{Abdomen}: We used the LITS \cite{5}, KITS\cite{6}, and PANCREAS \cite{29} datasets. LITS contains 131 cases, KITS has 210 cases, and PANCREAS includes 82 cases. These datasets are labeled with the liver, kidney, pancreas, and corresponding tumors; we only selected organs for training.

\textit{Pelvis}: We used the Word \cite{27} and CT-ORG \cite{28} datasets. Word contains 150 cases with 16 labeled organs; we selected the rectum, left femur, and right femur. CT-ORG includes 140 cases with 4 labeled organs (lungs, liver, kidneys, and bladder); we selected the bladder.

Since the head and neck datasets include labels for all selected organs, we utilized these datasets to analyze the quality of generated pseudo labels and for feature visualization.

\subsection{Experiment Setup}
\subsubsection{Implementation Details}
In pre-processing, we divided all 3D CT images into 2D slices and adjusted the intensity of the CT scans to filter out irrelevant regions. Our built models whose backbone network is 2D Res U-Net. We used the stochastic gradient descent (SGD) algorithm with Nesterov momentum (µ= 0.999) as the optimiser, the initial learning rate was set to 0.001 in both the first and second stages and decayed as training proceeded. Using Dice loss as the segmentation loss. All the experiments were performed on an NVIDIA RTX 4090.

\subsubsection{Evaluation Metrics}
Dice Similarity Coefficient (DSC) and Average Symmetric Surface Distance (ASSD) were used to evaluate the segmentation results. DSC calculates the overlap between prediction and GT, and ASSD evaluates the quality of segmented boundaries by calculating the average of all the surface distances between the predicted and true boundaries. It is important to note that in the experiments, the validation set was also a partially labeled dataset, so the average metrics for each organ were calculated based only on the dataset in which the organ was labeled.

\begin{table*}[t!]
\centering
\caption{The segmentation results of each comparison method in the head and neck. Red and blue represent the optimal and suboptimal results, respectively.}
\resizebox{\textwidth}{!}{  
\begin{tabular}{ccccccccccccccccc}
\hline
                                         & \multicolumn{2}{c}{\textbf{Brainstem}}                                       & \multicolumn{2}{c}{\textbf{Left optic nerve}}                                & \multicolumn{2}{c}{\textbf{Right optic nerve}}                               & \multicolumn{2}{c}{\textbf{Chiasm}}                                          & \multicolumn{2}{c}{\textbf{Left parotid gland}}                              & \multicolumn{2}{c}{\textbf{Right parotid gland}}                             & \multicolumn{2}{c}{\textbf{Mandible}}                                        & \multicolumn{2}{c}{\textbf{Average}}                                         \\ \cline{2-17} 
\multirow{-2}{*}{\textbf{Head and Neck}} & \textbf{DSC↑}                         & \textbf{ASSD↓}                       & \textbf{DSC↑}                         & \textbf{ASSD↓}                       & \textbf{DSC↑}                         & \textbf{ASSD↓}                       & \textbf{DSC↑}                         & \textbf{ASSD↓}                       & \textbf{DSC↑}                         & \textbf{ASSD↓}                       & \textbf{DSC↑}                         & \textbf{ASSD↓}                       & \textbf{DSC↑}                         & \textbf{ASSD↓}                       & \textbf{DSC↑}                         & \textbf{ASSD↓}                       \\ \hline
Multi-Net                                & 86.13                                 & 1.38                                 & \textcolor{blue}{ \textbf{76.68}} & 0.92                                 & \textcolor{blue}{ \textbf{77.17}} & 1.85                                 & 47.18                                 & \textcolor{blue}{ \textbf{1.76}} & \textcolor{blue}{ \textbf{84.47}} & 1.35                                 & 81.45                                 & 1.66                                 & 89.63                                 & 2.15                                 & \textcolor{blue}{ \textbf{77.53}} & 1.58                                 \\
TAL                                      & \textcolor{red}{ \textbf{87.61}} & \textcolor{red}{ \textbf{1.11}} & 74.57                                 & 1.51                                 & 71.08                                 & 1.47                                 & 42.18                                 & 1.86                                 & 83.04                                 & 1.30                                  & 81.58                                 & 1.70                                  & 90.67                                 & \textcolor{red}{ \textbf{1.35}} & 75.82                                 & 1.47                                 \\
ME                                       & 87.06                                 & 1.17                                 & 72.41                                 & 1.16                                 & 72.09                                 & 1.69                                 & 46.43                                 & 2.04                                 & 84.65                                 & 1.31                                 & 82.22                                 & 1.73                                 & 90.50                                  & \textcolor{blue}{ \textbf{1.58}} & 76.48                                 & 1.53                                 \\
DoDNet                                   & 87.18                                 & 26.03                                & 71.77                                 & 2.55                                 & 71.49                                 & 2.39                                 & 37.08                                 & 2.01                                 & 56.50                                 & 21.31                                & 56.25                                 & 22.49                                & \textcolor{blue}{ \textbf{90.83}} & 55.41                                & 67.30                                  & 18.88                                \\
CLIP-driven                              & 85.14                                 & 20.03                                & 63.51                                 & 3.01                                 & 62.72                                 & 5.14                                 & 35.14                                 & 3.01                                 & 55.14                                 & 25.14                                & 53.14                                 & 23.41                                & 87.14                                 & 35.17                                & 63.13                                 & 16.41                                \\
Co-training                              & 86.89                                 & 1.30                                  & 74.54                                 & 1.10                                  & 72.55                                 & \textcolor{blue}{ \textbf{1.30}}  & \textcolor{blue}{ \textbf{47.91}} & 2.46                                 & 85.20                                  & \textcolor{blue}{ \textbf{1.29}} & \textcolor{blue}{ \textbf{82.94}} & \textcolor{blue}{ \textbf{1.59}} & \textcolor{red}{ \textbf{90.96}} & 1.17                                 & 77.28                                 & \textcolor{blue}{ \textbf{1.46}} \\ \hline
\textbf{Ours}                            & \textcolor{blue}{ \textbf{87.29}} & \textcolor{blue}{ \textbf{1.15}} & \textcolor{red}{ \textbf{78.29}} & \textcolor{blue}{ \textbf{1.05}} & \textcolor{red}{ \textbf{79.45}} & \textcolor{red}{ \textbf{1.28}} & \textcolor{red}{ \textbf{50.22}} & \textcolor{red}{ \textbf{1.72}} & \textcolor{red}{ \textbf{85.44}} & \textcolor{red}{ \textbf{1.27}} & \textcolor{red}{ \textbf{83.66}} & \textcolor{red}{ \textbf{1.45}} & 90.65                                 & 1.65                                 & \textcolor{red}{ \textbf{79.29}} & \textcolor{red}{ \textbf{1.37}} \\ \hline
\end{tabular}
}  
\label{tab:1}
\end{table*}

\begin{figure*}[t!]  
\centering
  \includegraphics[width=0.7\textwidth]{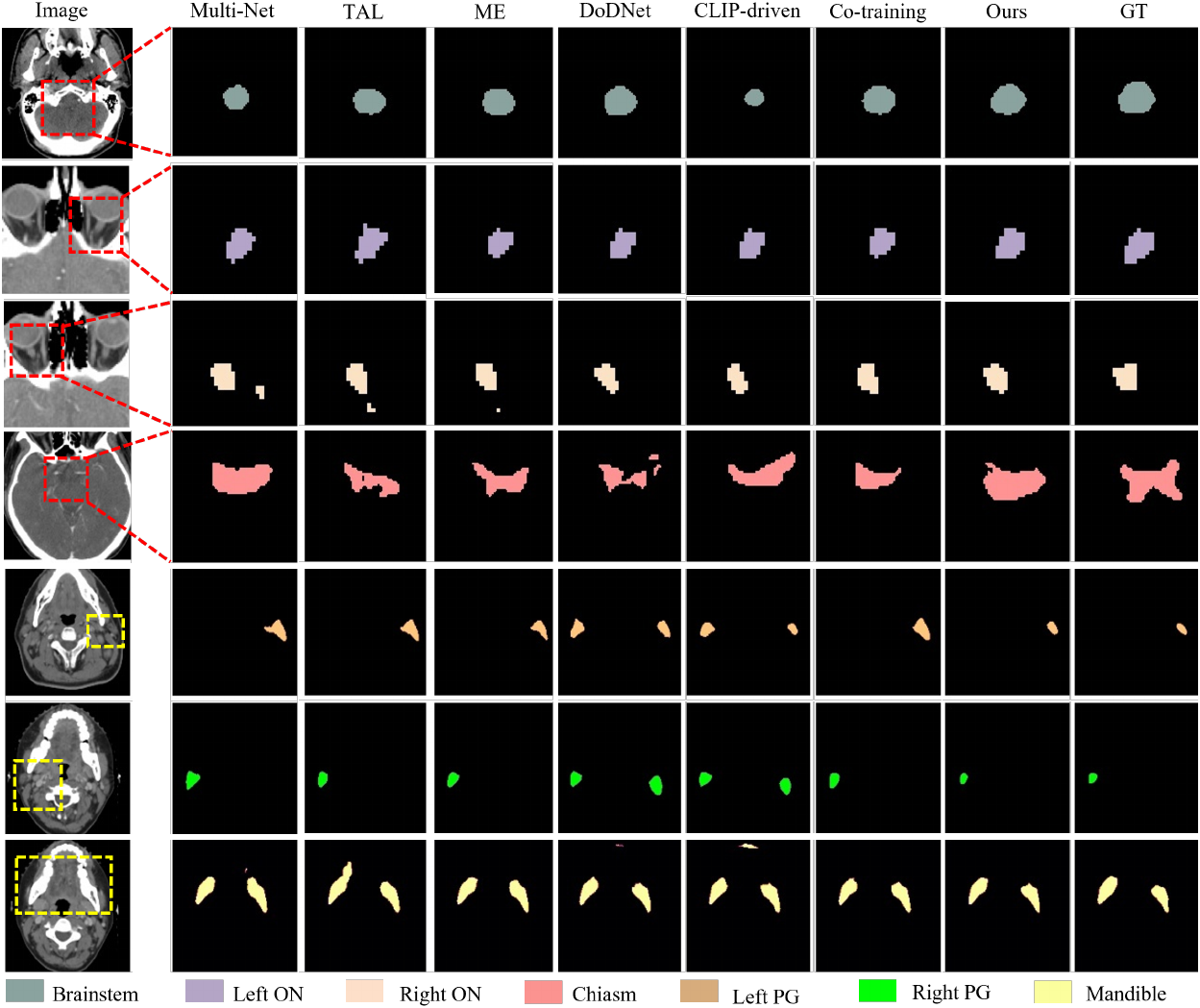}  
  \caption{Visualization of segmentation results for each method in head and neck. The red dashed box represents the selected and enlarged region, while the yellow dashed box only represents the selected region.}
  \label{fig:4}
\end{figure*}

\subsection{Comparison With State-of-the-Art Methods}

We compared our method with several existing methods: (1) Multi-Net: separate segmentation models for each dataset; (2) TAL \cite{10}: channel adjustment; (3) ME \cite{11}: marginal loss with channel adjustment; (4) DoDNet \cite{12}: Conditionally-guided approach; (5) CLIP-driven method \cite{14}: using "a CT of [organ]" prompt instead of DoDNet's one-hot encoding; (6) Co-training \cite{8}: two stage pseudo label-based approach. To ensure fairness, we used the same backbone architecture and training strategy for all methods, as detailed in Implementation Details.

\begin{figure*}[h]  
\centering
  \includegraphics[width=0.7\textwidth]{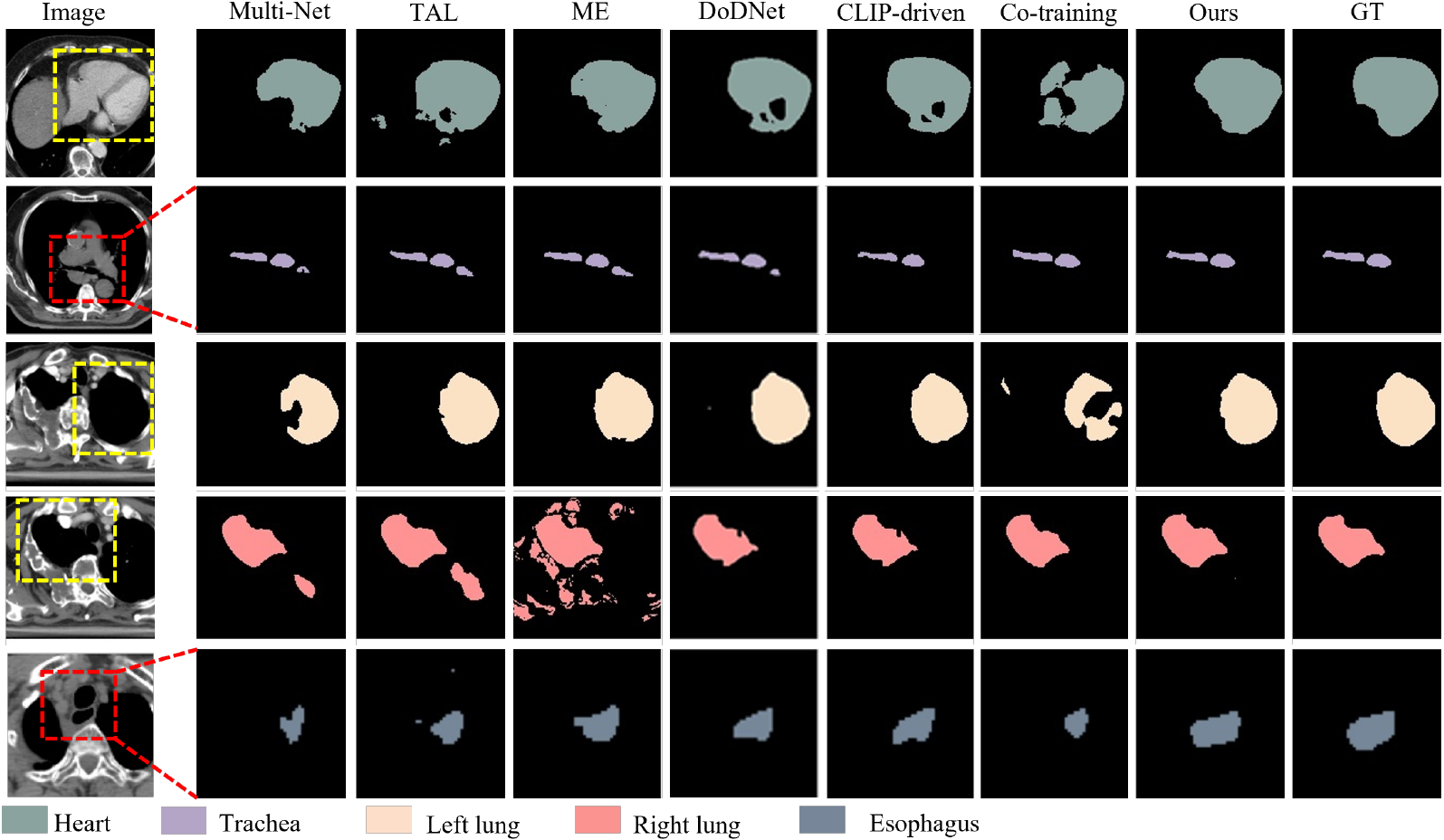}  
  \caption{Visualization of segmentation results for each method in chest. The red dashed box represents the selected and enlarged region, while the yellow dashed box only represents the selected region.}
  \label{fig:5}
\end{figure*}

\begin{figure*}[t!]  
\centering
  \includegraphics[width=0.7\textwidth]{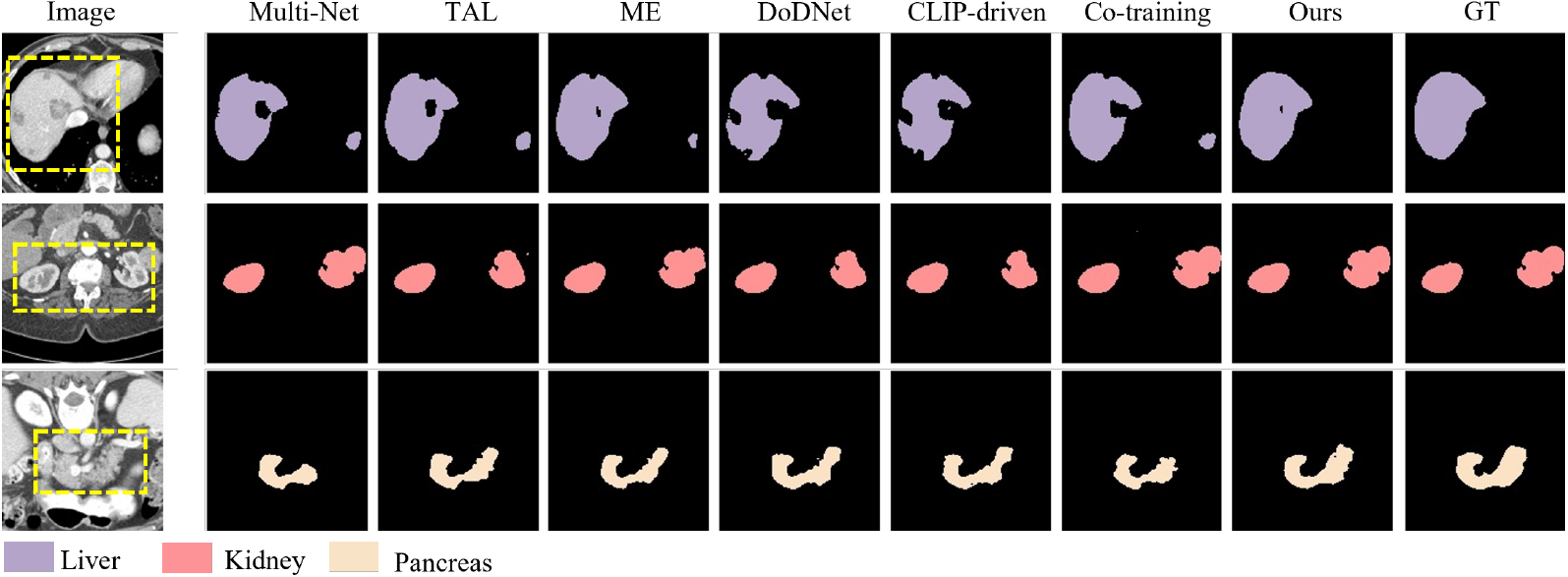}  
  \caption{Visualization of segmentation results for each method in abdomen. The red dashed box represents the selected and enlarged region, while the yellow dashed box only represents the selected region.}
  \label{fig:6}
\end{figure*}

\begin{figure*}[t!]  
\centering
  \includegraphics[width=0.7\textwidth]{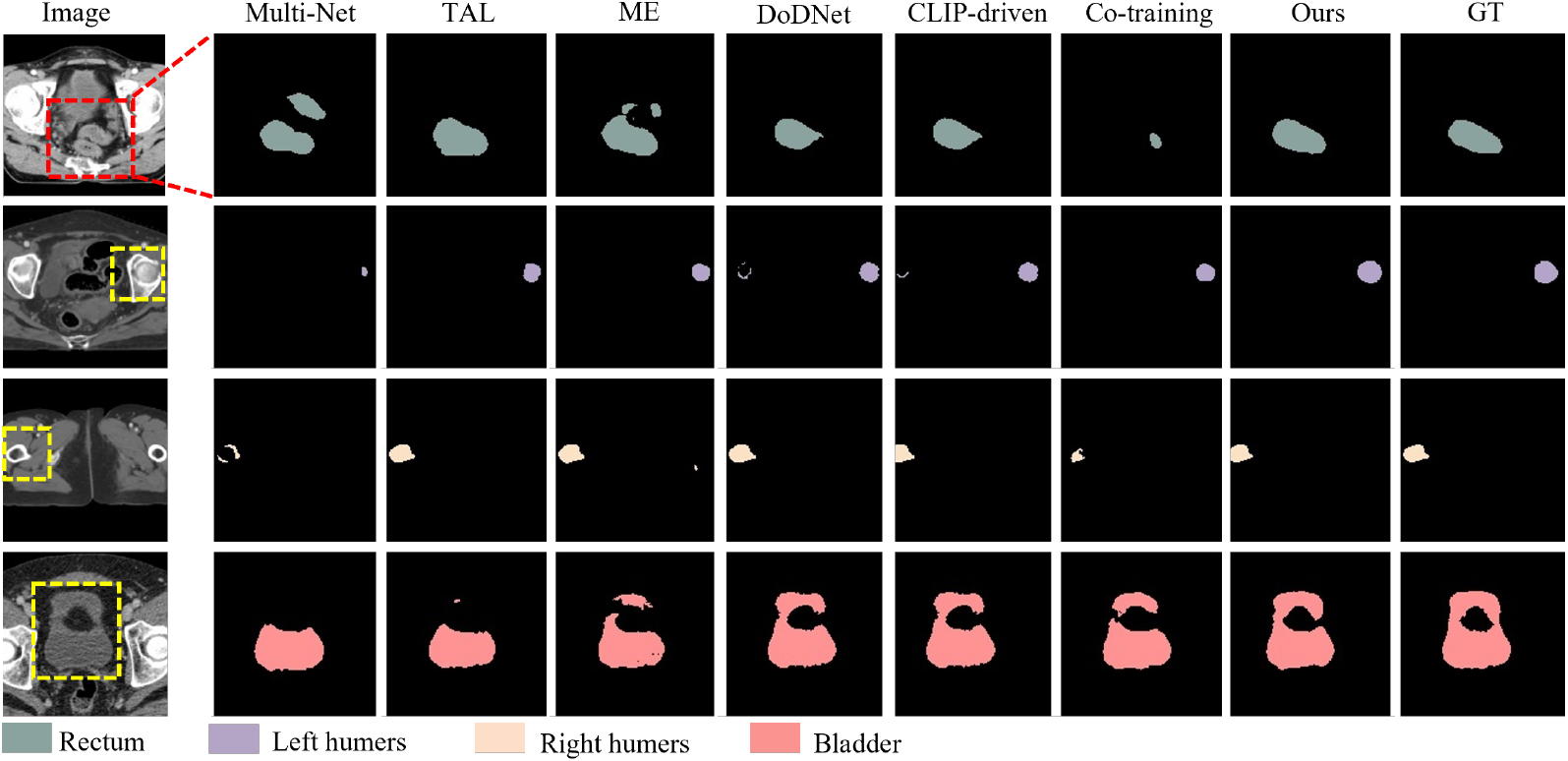}  
  \caption{Visualization of segmentation results for each method in pelvis. The red dashed box represents the selected and enlarged region, while the yellow dashed box only represents the selected region.}
  \label{fig:7}
\end{figure*}

Tables ~\ref{tab:1}, ~\ref{tab:2}, ~\ref{tab:3} and ~\ref{tab:4} present the segmentation results for the head and neck, chest, abdomen, and pelvis, respectively. Figs.~\ref{fig:4}, ~\ref{fig:5}, ~\ref{fig:6} and ~\ref{fig:7} provide visualizations of these methods. The following conclusions can be drawn from these results:

Unified segmentation models generally outperform Multi-Net, as indicated by combined DSC and ASSD metrics. Among channel adjustment methods, ME outperforms TAL by leveraging non-overlapping organ annotations across datasets. The conditionally-guided DoDNet achieves sub-optimal results in the chest and abdomen but performs poorly in the head and neck and pelvis, particularly for the chiasm and parotid glands; The method also struggles to distinguish between symmetrical structures (e.g., parotid glands and humerus), as noted in COSST \cite{7}. The CLIP-driven method performs poorly across all regions, especially for less frequent organs and structures like the chiasm.
The Co-training method, based on two stage pseudo-labeling, achieves competitive results, particularly in the pelvis. Overall, our method outperforms others across all regions, especially for small organs like the chiasm and elongated organs like the esophagus. Visually, our method aligns more closely with GT, avoiding the segmentation errors of channel adjustment and the issues with symmetric structures in conditional guidance methods.
\begin{table*}[t!]
\centering
\caption{The segmentation results of each comparison method in the chest. Red and blue represent the optimal and suboptimal results, respectively.}
\resizebox{0.8\textwidth}{!}{  
\begin{tabular}{ccccccccccccc}
\hline
                                 & \multicolumn{2}{c}{\textbf{Heart}}                                           & \multicolumn{2}{c}{\textbf{Trachea}}                                         & \multicolumn{2}{c}{\textbf{Left Lung}}                                       & \multicolumn{2}{c}{\textbf{Right Lung}}                                      & \multicolumn{2}{c}{\textbf{Esophagus}}                                       & \multicolumn{2}{c}{\textbf{Average}}                                         \\ \cline{2-13} 
\multirow{-2}{*}{\textbf{Chest}} & \textbf{DSC↑}                          & \textbf{ASSD↓}                        & \textbf{DSC↑}                          & \textbf{ASSD↓}                        & \textbf{DSC↑}                          & \textbf{ASSD↓}                        & \textbf{DSC↑}                          & \textbf{ASSD↓}                        & \textbf{DSC↑}                          & \textbf{ASSD↓}                        & \textbf{DSC↑}                          & \textbf{ASSD↓}                        \\ \hline
Multi-Net                        & 91.07                                 & 6.31                                 & 91.71                                 & 0.88                                 & 92.03                                 & 4.67                                 & 96.06                                 & 2.03                                 & 72.90                                  & 1.08                                 & 88.75                                 & 2.99                                 \\
TAL                              & 89.99                                 & 6.04                                 & 90.46                                 & \textcolor{red}{ \textbf{0.54}} & 95.04                                 & \textcolor{blue}{ \textbf{1.55}} & 96.29                                 & \textcolor{blue}{ \textbf{1.35}} & 69.47                                 & 1.07                                 & 88.24                                 & 2.11                                 \\
ME                               & \textcolor{blue}{ \textbf{92.33}} & \textcolor{blue}{ \textbf{4.57}} & 89.44                                 & 1.29                                 & \textcolor{red}{ \textbf{95.41}} & \textcolor{red}{ \textbf{1.47}} & \textcolor{blue}{ \textbf{96.66}} & 1.68                                 & 69.47                                 & 1.11                                 & 88.66                                 & \textcolor{blue}{ \textbf{2.02}} \\
DoDNet                           & 91.84                                 & 44.94                                & \textcolor{blue}{\textbf{92.17}} & 66.66                                & 94.97                                 & 110.90                                & 96.12                                 & 112.87                               & \textcolor{blue}{ \textbf{74.84}} & 120.90                                & \textcolor{blue}{ \textbf{89.46}} & 91.25                                \\
CLIP-driven                      & 85.14                                 & 35.14                                & 86.14                                 & 55.14                                & 92.14                                 & 100.14                               & 91.47                                 & 90.87                                & 55.41                                 & 110.70                                & 82.06                                 & 78.40                                 \\
Co-training                      & 91.23                                 & 4.82                                 & 89.55                                 & 1.47                                 & 93.00                                    & 3.10                                  & 93.58                                 & 2.31                                 & 75.20                                  & \textcolor{red}{ \textbf{0.97}} & 88.78                                 & 2.53                                 \\ \hline
\textbf{Ours}                             & \textcolor{red}{ \textbf{93.18}} & \textcolor{red}{ \textbf{4.13}} & \textcolor{red}{ \textbf{92.78}} & \textcolor{blue}{ \textbf{0.63}} & \textcolor{blue}{ \textbf{95.30}}  & \textcolor{red}{ \textbf{1.47}} & \textcolor{red}{ \textbf{96.68}} & \textcolor{red}{ \textbf{1.20}}  & \textcolor{red}{ \textbf{76.73}} & \textcolor{blue}{ \textbf{0.99}} & \textcolor{red}{ \textbf{90.94}} & \textcolor{red}{ \textbf{1.68}} \\ \hline
\end{tabular}
}  
\label{tab:2}
\end{table*}

\begin{table}[h!]
\centering
\caption{The segmentation results of each comparison method in the abdomen. Red and blue represent the optimal and suboptimal results, respectively.}
\resizebox{0.5\textwidth}{!}{  
\begin{tabular}{ccccccccc}
\hline
                                   & \multicolumn{2}{c}{\textbf{Liver}}                                           & \multicolumn{2}{c}{\textbf{Kidney}}                                          & \multicolumn{2}{c}{\textbf{Pancreas}}                                        & \multicolumn{2}{c}{\textbf{Average}}                                         \\ \cline{2-9} 
\multirow{-2}{*}{\textbf{Abdomen}} & \textbf{DSC↑}                          & \textbf{ASSD↓}                        & \textbf{DSC↑}                          & \textbf{ASSD↓}                        & \textbf{DSC↑}                          & \textbf{ASSD↓}                        & \textbf{DSC↑}                          & \textbf{ASSD↓}                        \\ \hline
Multi-Net                          & \textcolor{red}{ \textbf{95.27}} & 10.29                                & \textcolor{blue}{ \textbf{95.61}} & 5.72                                 & 78.82                                 & 5.22                                 & 89.90                                  & 7.08                                 \\
TAL                                & 94.75                                 & \textcolor{blue}{ \textbf{8.41}} & 95.14                                 & \textcolor{red}{ \textbf{4.33}} & 79.23                                 & 4.37                                 & 89.71                                 & 5.70                                  \\
ME                                 & 95.03                                 & \textcolor{red}{ \textbf{7.90}}  & 94.60                                  & 5.35                                 & 79.59                                 & \textcolor{blue}{ \textbf{4.05}} & 89.74                                 & \textcolor{blue}{ \textbf{5.66}} \\
DoDNet                             & \textcolor{blue}{ \textbf{95.21}} & 15.88                                & 94.32                                 & 14.21                                & \textcolor{red}{ \textbf{82.01}} & 25.68                                & \textcolor{blue}{ \textbf{90.51}} & 10.11                                \\
CLIP-driven                        & 92.14                                 & 12.77                                & 93.16                                 & 15.14                                & 77.25                                 & 26.12                                & 87.52                                 & 18.01                                \\
Co-training                        & 94.63                                 & 17.74                                & 94.57                                 & 7.19                                 & 76.03                                 & 5.49                                 & 88.41                                 & 10.14                                \\ \hline
\textbf{Ours}                      & 95.08                                 & 8.65                                 & \textcolor{red}{ \textbf{96.66}} & \textcolor{blue}{ \textbf{4.58}} & \textcolor{blue}{ \textbf{80.49}} & \textcolor{red}{ \textbf{3.44}} & \textcolor{red}{ \textbf{90.74}} & \textcolor{red}{ \textbf{5.56}} \\ \hline
\end{tabular}
}  
\label{tab:3}
\end{table}

\begin{table}[h!]
\centering
\caption{The segmentation results of each comparison method in the pelvis. Red and blue represent the optimal and suboptimal results, respectively.}
\resizebox{0.5\textwidth}{!}{  
\begin{tabular}{ccccccccccc}
\hline
                                         & \multicolumn{2}{c}{\textbf{Rectum}}                                          & \multicolumn{2}{c}{\textbf{Left humers}}                                     & \multicolumn{2}{c}{\textbf{Right humers}}                                    & \multicolumn{2}{c}{\textbf{Bladder}}                                         & \multicolumn{2}{c}{\textbf{Average}}                                         \\ \cline{2-11} 
\multirow{-2}{*}{\textbf{Pelvis}} & \textbf{DSC↑}                          & \textbf{ASSD↓}                        & \textbf{DSC↑}                          & \textbf{ASSD↓}                        & \textbf{DSC↑}                          & \textbf{ASSD↓}                        & \textbf{DSC↑}                          & \textbf{ASSD↓}                        & \textbf{DSC↑}                          & \textbf{ASSD↓}                        \\ \hline
Multi-Net                                & 79.01                                 & 2.30                                  & 93.70                                  & 1.49                                 & 92.97                                 & 1.21                                 & 89.59                                 & 15.26                                & 88.82                                 & 5.24                                 \\
TAL                                      & 80.17                                 & 2.38                                 & 93.50                                  & \textcolor{red}{ \textbf{0.94}} & 92.57                                 & \textcolor{blue}{ \textbf{0.93}} & \textcolor{blue}{ \textbf{90.90}}  & \textcolor{red}{ \textbf{7.15}} & 89.29                                 & \textcolor{red}{ \textbf{2.85}} \\
ME                                       & \textcolor{blue}{ \textbf{81.99}} & \textcolor{blue}{ \textbf{2.06}} & \textcolor{blue}{ \textbf{93.86}} & \textcolor{blue}{ \textbf{1.08}} & 92.86                                 & 1.13                                 & 90.37                                 & 9.14                                 & 89.77                                 & 3.35                                 \\
DoDNet                                   & 76.04                                 & 30.93                                & 88.73                                 & 32.95                                & 90.98                                 & 31.24                                & 70.24                                 & 24.91                                & 81.50                                  & 30.01                                \\
CLIP-driven                              & 65.14                                 & 35.41                                & 76.14                                 & 45.12                                & 75.96                                 & 35.44                                & 65.41                                 & 32.41                                & 70.66                                 & 37.10                                 \\
Co-training                              & 81.97                                 & 2.30                                  & 93.80                                  & 1.14                                 & \textcolor{blue}{ \textbf{93.52}} & 1.73                                 & 90.69                                 & 9.51                                 & \textcolor{blue}{ \textbf{89.99}} & 3.67                                 \\ \hline
\textbf{Ours}                            & \textcolor{red}{ \textbf{83.79}} & \textcolor{red}{ \textbf{1.73}} & \textcolor{red}{ \textbf{94.47}} & 1.01                                 & \textcolor{red}{ \textbf{94.57}} & \textcolor{red}{ \textbf{0.82}} & \textcolor{red}{ \textbf{91.14}} & \textcolor{blue}{ \textbf{8.88}} & \textcolor{red}{ \textbf{90.99}} & \textcolor{blue}{ \textbf{3.11}} \\ \hline
\end{tabular}
}  
\label{tab:4}
\end{table}

\subsection{Ablation Studies}
\subsubsection{Effectiveness of Difference Mutual Learning}
The difference learning introduced in the first stage can enhance the model's ability to segment the current organ, improve the distinctiveness of different student models, and enhance the quality of the generated pseudo labels. To evaluate its effectiveness, we conducted extensive ablation studies and analyzed the following aspects:

\noindent\textit{\textbf{Metric.}} Table ~\ref{tab:5} compares mean DSC across models. After introducing the label-level difference mutual learning loss (PD), the DSC for all body parts were improved, especially with the DSC of chest increasing to 90.05 (an increase of 1.06). The addition of feature-level difference loss (FD) further improved the average DSC to 78.62 (head and neck), 90.07 (chest), 90.43 (abdomen), and 89.37 (pelvis), demonstrating that incorporating information from other datasets can enhance the segmentation performance of the current model.

\begin{table}[t]
\centering
\caption{The ablation study result of difference mutual learning. The data in the table are the average DSC of the segmented organs of each body region.}
\resizebox{0.5\textwidth}{!}{  
\begin{tabular}{ccccccc}
\hline
\textbf{Baseline} & \textbf{PD} & \textbf{FD} & \textbf{Head and Neck}  &\textbf{Chest}  &\textbf{Abdomen}  &\textbf{Pelvis}\\ \hline
\checkmark            &\phantom      & \phantom                  & 77.54                  & 88.99          & 89.88            & 88.80       \\
\checkmark        &\checkmark       &\phantom          & 77.74                  & 89.51          & 90.38            & 89.28     \\ \hline
\checkmark        &\checkmark   &\checkmark   & \textbf{78.62}         & \textbf{90.07} & \textbf{90.43}   & \textbf{89.37}    \\ \hline
\end{tabular}
}  
\label{tab:5}
\end{table}

\begin{figure}[t!]  
\centering
  \includegraphics[width=0.5\textwidth]{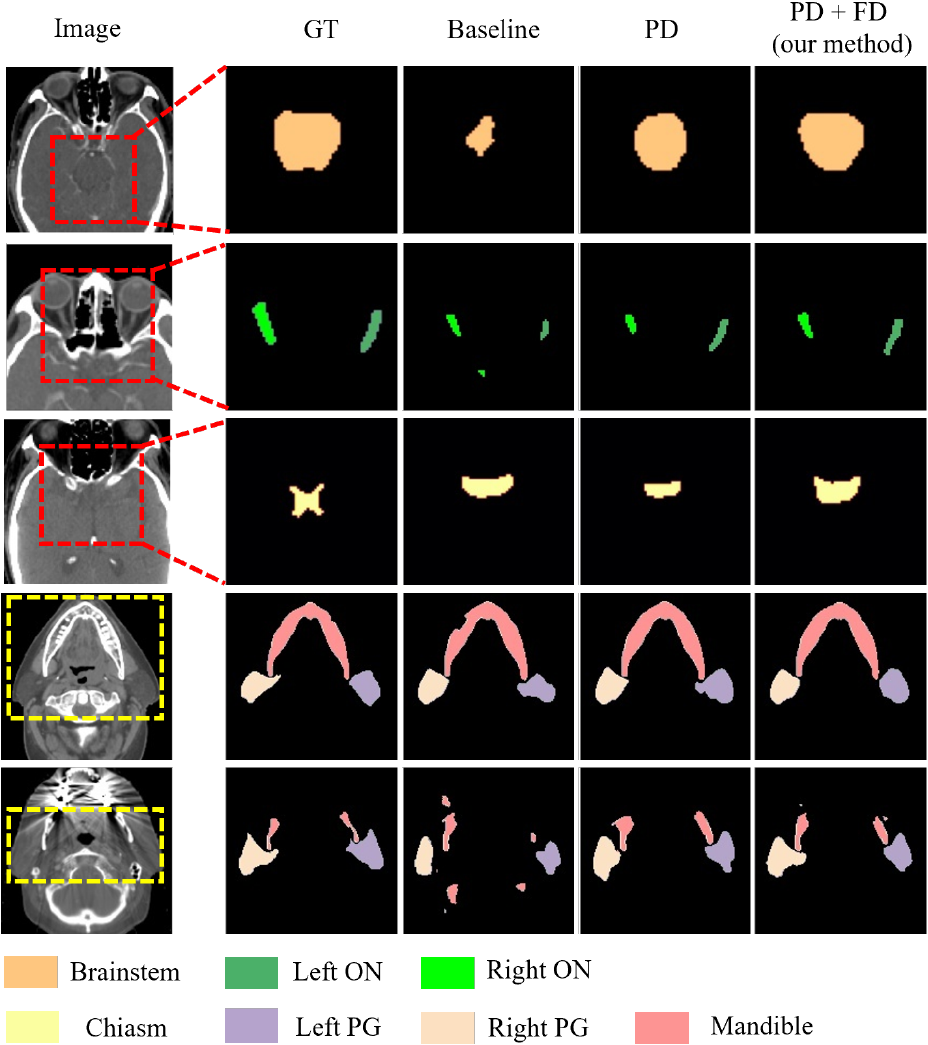}  
  \caption{Visualization of the generated pseudo labels after the first stage. The red dashed box represents the selected and enlarged region, while the yellow dashed box only represents the selected region.}
  \label{fig:8}
\end{figure}

\noindent\textit{\textbf{Pseudo-Label Analysis.}} To verify that the introduction of Prediction-level Difference (PD) and Feature-level Difference (FD) can produce higher-quality pseudo labels, we took the head and neck PDDCA and StructSeg datasets as examples. The models trained on these datasets are referred to as \( P_1 \) and \( P_2 \) (partial-organ segmentation model), respectively. \( P_1 \) generated pseudo labels for the brainstem and left and right optic nerves on the StructSeg dataset, while \( P_2 \) generated pseudo labels for the chiasm, left and right parotid glands, and the mandible on the PDDCA dataset, and then compared these with the true labels of the two datasets. The DSC between the pseudo labels and the true labels showed that, without the introduction of difference learning, the average DSC for the seven organs was 56.67 (53.95 for PDDCA, 58.70 for StructSeg). Introducing PD loss increased the average DSC to 57.64, with a particularly significant improvement in the PDDCA dataset, rising to 57.17 (an increase of 3.22). Further introducing feature difference loss (PD + FD) improved the average DSC to 58.44, with PDDCA rising to 57.87 and StructSeg to 58.88. Fig. ~\ref{fig:8} also shows that the pseudo labels generated with label and feature difference losses more closely resemble the true labels, especially for organs such as the brainstem and mandible, indicating that the introduced difference learning enable the model to perceive the presence of other organs, thereby generating higher quality pseudo labels on other datasets.

\begin{figure}[t!]  
\centering
  \includegraphics[width=0.5\textwidth]{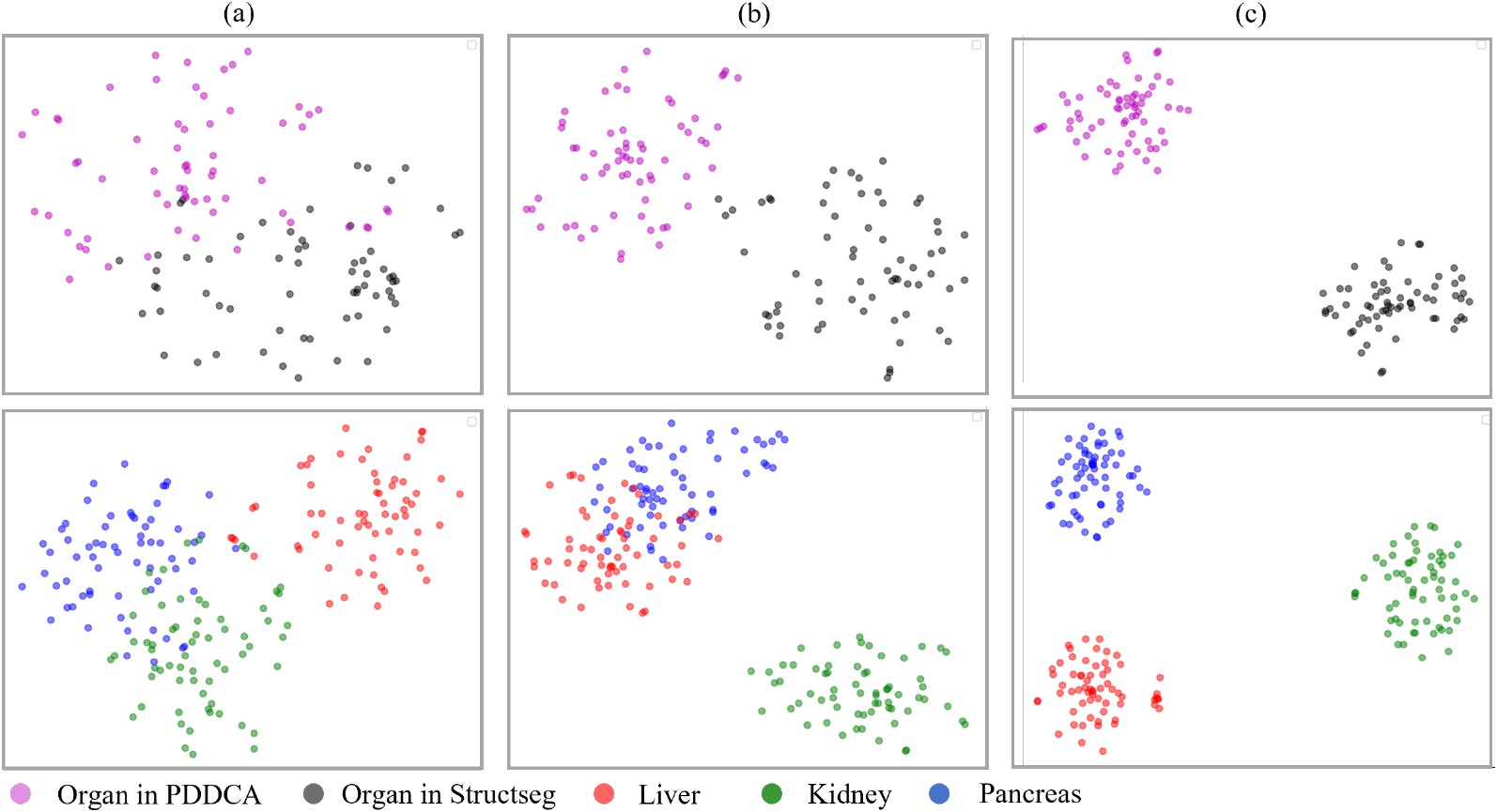}  
  \caption{Feature visualization results. (a) no difference loss was introduced. (b) Introduction of PD. (c) Reintroduce FD.}
  \label{fig:9}
\end{figure}

\noindent\textit{\textbf{Feature Visualization.}}
To verify that the difference learning can make the features of different organs extracted by different models more distinguishable, we used t-SNE to visualize the high-dimensional features extracted by different models on the same dataset. As shown in Fig.~\ref{fig:9}, without difference learning (Fig.~\ref{fig:9} (a)), the features of different organs significantly overlap, which reduces the segmentation performance of the model and the quality of the pseudo labels. Introducing PD loss (Fig.~\ref{fig:9} (b)) provides some distinguishability among the features of different organs, but overlap still persists. After adding FD loss (Fig.~\ref{fig:9} (c)), the features of different organs are clearly separated, leading to higher precision in segmenting different organs by different models, and also improving the quality of pseudo labels generated on other datasets, significantly reducing the occurrence of overlap with labels from other organs.

\subsubsection{Effectiveness of Similarity Mutual Learning}

In the second stage, we conducted an ablation study to verify the effectiveness of the proposed similarity learning, as shown in Table ~\ref{tab:6}. The baseline model was trained under the supervision of combined labels, achieving an average DSC of 77.34 (head and neck), 88.72 (chest), 88.63 (abdomen), and 90.30 (pelvis). The introduction of LS improved the average DSC, especially for the abdomen, increasing by 1.37. Further addition of the FS (especially DFS) brought the average DSC to 79.29 (head and neck), 90.94 (chest), 90.74 (abdomen), and 90.99 (pelvis). These results confirm that similarity learning can fully utilize the true labels of other datasets, increase supervisory information, and enhance the performance of multi-organ segmentation.


\begin{table}[t!]
\centering
\caption{The ablation study result of similarity mutual learning. The data in the table are the average DSC of the segmented organs of each body region.}
\resizebox{0.5\textwidth}{!}{  
\begin{tabular}{cccccccc}
\hline
\textbf{Baseline} & \textbf{PS} & \textbf{FS}& \textbf{DFS} & \textbf{Head and Neck}  &\textbf{Chest}  &\textbf{Abdomen}  &\textbf{Pelvis}\\ \hline
\checkmark            &\phantom     & \phantom   & \phantom               & 77.34                  & 88.72          & 88.63            & 90.30       \\
\checkmark        &\checkmark       &\phantom     &\phantom    & 78.16                  & 89.77          & 90.00               & 90.60      \\
\checkmark        &\checkmark   &\checkmark  &\phantom  & 77.92                  & 89.84          & 90.21            & 90.70    \\\hline
\checkmark        &\checkmark  &\phantom  &\checkmark   & \textbf{79.29}         & \textbf{90.94} & \textbf{90.74}   & \textbf{90.99}    \\ \hline
\end{tabular}
}  
\label{tab:6}
\end{table}

\subsubsection{Effectiveness of DFS}

We also evaluated dynamic feature similarity mutual learning in the second stage (Table ~\ref{tab:6}). Comparing with static feature similarity mutual learning, our proposed DFS outperformed it across all regions: it improved mean DSC by 1.37 (head and neck), 1.1 (chest), 0.53 (abdomen), and 0.29 (pelvis). This demonstrates that DFS effectively transfers correct knowledge, enhancing model supervision and performance.

\section{Discussion}
In this paper, we have proposed a two-stage mutual learning approach to utilize partially labeled datasets. The mutual learning between models for segmenting different organs in the first stage not only improves each model's ability to segment the labeled organs, but also enhances its perception of unlabeled organs to generate higher quality pseudo labels. The second stage is to train models to learn from each other with fully labeled datasets containing pseudo labels. The supervised information includes the true labels of different datasets as well as the pseudo labels generated after the first stage, while the features extracted by different models can also be dynamically transferred to each other to achieve mutual enhancement between models, thus improving the performance of multi-organ segmentation models.

The effectiveness of our method has been demonstrated through the experiments on diverse datasets encompassing the head and neck, chest, abdomen, and pelvis, which has consistently achieved superior performance in each of these regions, surpassing the state-of-the-art methods (see Tables ~\ref{tab:1}, ~\ref{tab:2}, ~\ref{tab:3} and ~\ref{tab:4}). Additionally, visual results show that our method's segmentation results closely matches the ground truths (see Figs.~\ref{fig:4}, ~\ref{fig:5}, ~\ref{fig:6} and ~\ref{fig:7}). According to the results obtained by different methods, incorporating organ-specific priors, as evidenced in TAL \cite{10} and ME \cite{11}, and employing pseudo-labelling through Co-training \cite{8}, effectively enriches the supervisory signals, thereby enhancing segmentation outcomes. Conditional information-guided methods excel with specific organs but struggle with smaller structures, and they are unable to differentiate between symmetric structures, such as the left and right parotid glands and the left and right humers, as shown in the fifth column of Figs. ~\ref{fig:4} and ~\ref{fig:7}. Although CLIP-driven method has achieved significant success in the segmentation of abdominal organs\cite{14}, it relies on large datasets for training, and most of the images pre-trained by CLIP are natural images. Therefore, further exploration is needed to adapt this method to medical imaging.

In medical images, organ sizes and locations are relatively fixed, serving as valuable prior information for multi-organ segmentation tasks. Despite variations in labeled organs across different datasets, the size and location information is crucial for multi-organ segmentation. Previous methods leveraged this information to regulate predictions of unlabeled organs or used organ size and location as priors to improve performance over independent training\cite{7,9,11}. However, these methods often overlook the richer feature-level information extracted by different models. The first stage of our model integrates both label and feature-level mutual difference learning, enhancing segmentation accuracy for labeled organs and improving the reliability of pseudo labels for unlabeled organs in other datasets.

After generating pseudo labels in the first stage, each dataset contains true labels and pseudo labels. pseudo labels can enhance supervisory information, and previous work \cite{7,8,9} has shown that training with pseudo labels can yield results comparable to or better than independent training. However, the presence of pseudo labels makes it difficult to further improve segmentation accuracy. Previous methods include co-training networks to update pseudo labels \cite{8}, the introduction of organ priors \cite{9}, and pseudo-label filtering mechanisms \cite{7}. Our method differs by fully integrating label and feature information across datasets. At the label level, we use true labels from other datasets to assist in training; at the feature level, we introduce a dynamic feature mutual learning mechanism that allows models to exchange accurate feature information. As a result, compared to previous methods, our method supervises with more information, including true and pseudo labels as well as correct features, thus achieving superior performance.

The proposed mutual learning strategy exhibits significant potential in multi-organ segmentation across diverse anatomical regions, providing novel insights for the tasks of medical image processing, such as imaging diagnosis and classification.

Our method still has limitations. First, training several models concurrently is a demanding task. In the future, more straightforward and efficient training methods will be designed. Second, The concept of mutual learning has inspired us to believe that incorporating datasets from multiple anatomical regions in training can potentially improve the accuracy even further. The future work will focus on leveraging abdominal parameters to optimize the segmentation of the organs in the head and neck. Lastly, the paucity of public datasets for regions such as the head and neck restricts our method's scope, indicating a need for expanded research.

\section{Conclusion}
In this study, we propose a two-stage multi-organ segmentation method based on the idea of mutual learning, which can maximise the use of label information in each dataset and prompt beneficial information exchange between different student models. By performing difference mutual learning as well as similarity mutual learning for multiple models in two stages respectively, the optimal model that can segment all organs at once without additional post-processing steps is obtained. The experimental results show that our method outperforms previous approaches on nine publicly available datasets containing the head and neck, chest, abdomen and pelvis. In addition, the ablation study also validates the effectiveness of each module proposed in the study.

\bibliographystyle{IEEEtran}
\bibliography{ref}

\end{document}